\pdfoutput=1

\documentclass[11pt]{article}

\usepackage[final]{acl}

\usepackage{times}
\usepackage{latexsym}
\usepackage{url}
\usepackage{tabularx}
\usepackage{multirow}

\usepackage[T1]{fontenc}

\usepackage[utf8]{inputenc}

\usepackage{color}

\usepackage{microtype}

\usepackage{inconsolata}

\usepackage{graphicx}

\usepackage{listings}
\lstdefinestyle{mystyle}{
    backgroundcolor=\color{backcolour},   
    commentstyle=\color{codegreen},
    keywordstyle=\color{magenta},
    numberstyle=\tiny\color{codegray},
    stringstyle=\color{codepurple},
    basicstyle=\ttfamily\footnotesize,
    breakatwhitespace=false,         
    breaklines=true,                 
    captionpos=b,                    
    keepspaces=true,                 
    numbers=left,
    numbersep=5pt,                  
    showspaces=false,                
    showstringspaces=false,
    showtabs=false,                  
    tabsize=2
}

\newcommand{\cready}[1]{{}}

\title{Naturally Occurring Feedback is Common, Extractable and Useful}

\author{
 \textbf{Shachar Don-Yehiya\textsuperscript{1}} \qquad
 \textbf{Leshem Choshen \textsuperscript{2,3}} \qquad
 \textbf{Omri Abend\textsuperscript{1}} \\
 \textsuperscript{1}The Hebrew University of Jerusalem,
 \textsuperscript{2}MIT,
 \textsuperscript{3}MIT-IBM Watson AI Lab \\
   \texttt{\{first.last\}@mail.huji.ac.il}
}

\begin{document}

\maketitle

\begin{abstract}
Human feedback data is a critical component in developing language models. However, collecting this feedback is costly and ultimately not scalable. 
Inspired by the way human interlocutors provide spontaneous unsolicited feedback to each other, we propose to extract 
feedback that users naturally include when interacting with chat models.
We manually annotated conversations to confirm the presence of naturally occurring feedback in a standard corpus, finding that as much as $\sim$30\% of the chats include explicit feedback.
Comparing to older datasets, we find that naturally occurring feedback is more prevalent in recent conversation datasets, suggesting that more than ever, naturally occurring feedback can serve as a valuable resource for feedback data.
We propose a method for automatically extracting this feedback, and apply it to over 1M conversations to obtain hundreds of thousands of feedback samples. 
The extracted feedback shows promise: training with it improves over baseline models and enhances model alignment to human preferences.
\end{abstract}


\begin{figure}[t]
\includegraphics[width=\columnwidth]{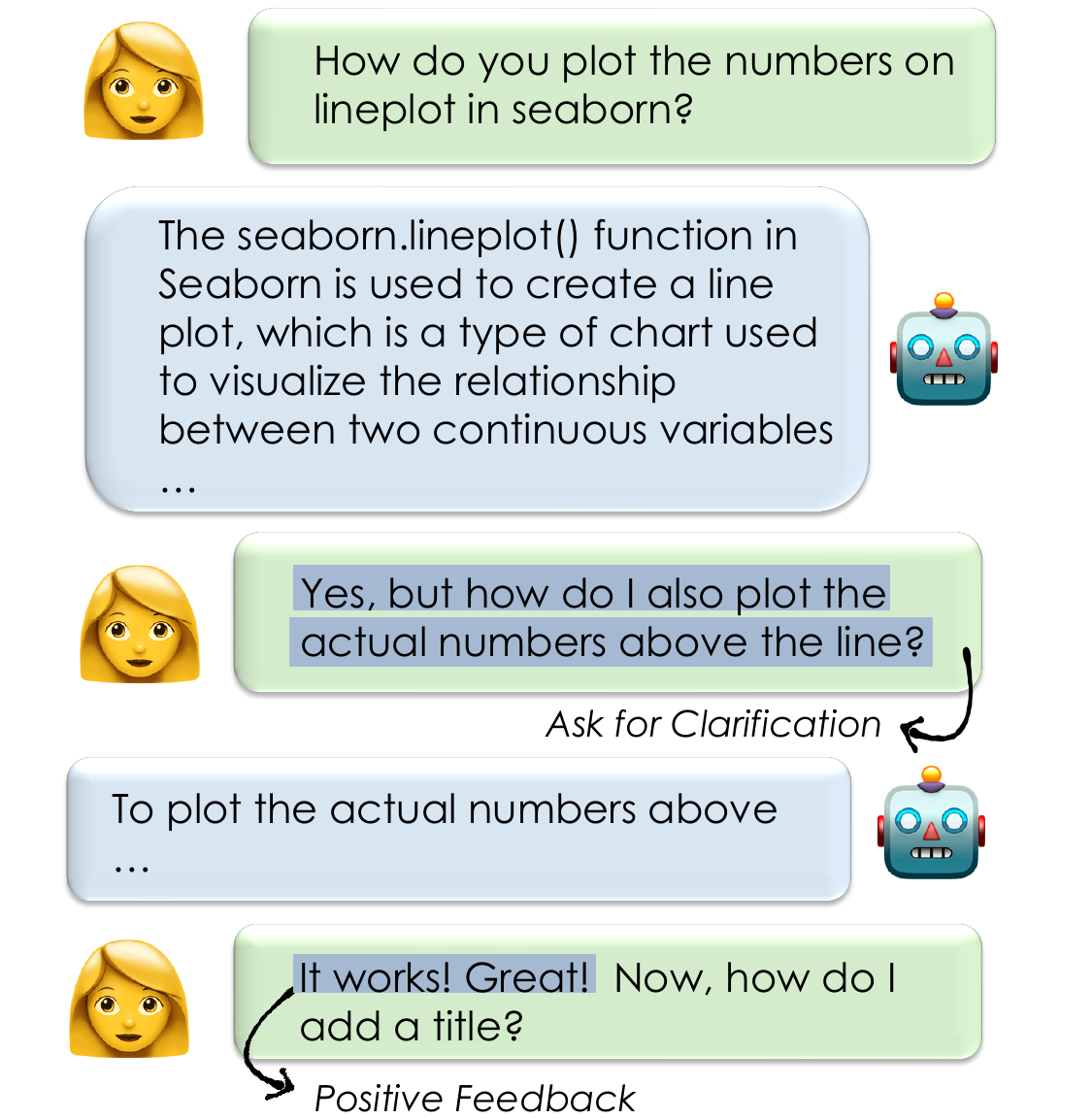}
\caption{Naturally occurring feedback example. 
         The feedback contained in the user responses is highlighted.}
\label{fig:fig_1}
\end{figure}

\section{Introduction}

Human feedback is a valuable resource for large language model (LLM) development. The current standard model training process includes a pretraining phase \citep{radford2019language}, 
followed by an alignment phase, where the model is usually fine-tuned and trained with reinforcement learning on human preference data \citep{bai2022training, touvron2023llama}. The more data at hand, the better the model \citep{kaplan2020scaling, roberts2023scaling}.
However, collecting such data usually requires costly human labor, limiting its scalability. 

Humans nevertheless do not need commentators to know that their conversation partner is satisfied. Rather, they infer it from the communication itself.
We suggest employing a similar rationale with LLMs, and consider conversations as a source for natural human feedback (see Fig.~\ref{fig:fig_1}). 
As naturalistic feedback comes in diverse forms and even implicit forms 
(e.g., the user continues to the next question feeling satisfied/displeased with the model's response), in this work we focus on explicit cues, such as when the user directly refers to the quality of the model's response (e.g.,
\textit{``thank you!''}, or \textit{``that's wrong''}) or rephrases and asks the same question again (cf. \S\ref{sec:feedback_taxonomy}). 

One advantage of this approach is that this form of feedback is potentially closer to the feedback given by two human interlocutors \citep{bassiri2011interactional, 5a61a38b-79fc-30d4-a76e-8a66a7707021}, possibly containing genuine information for better alignment.

With the introduction of general assistant models like ChatGPT \citep{openai2024gpt4} and OpenAsistant \citep{kopf2024openassistant}, human-model interactions have become very prominent, not only among machine learning experts but also among the general public. Thus,  huge amounts of conversation data are potentially available \citep{donyehiya2024sharelmcollectionplugincontributing}, and those can be viewed as potential feedback data.

We manually annotate and show that naturally occurring feedback is indeed prevalent in conversation data, finding $101$ feedback cases within the $300$ conversations we examined (\S\ref{sec:manual_annotations}).
Furthermore, we find that naturally occurring feedback is more common in recently collected data than in older data, possibly due to users raising their expectations and being able to conduct a more ``human-like'' conversation with the model (\S\ref{sec:up-to-date-feedback}). 
This further underscores the importance of ever-growing data resources, over static datasets. Models keep improving and therefore the data used to align them should evolve too \citep{openhumanfeedback}. 

We introduce a method to automatically extract the naturally occurring feedback from human-model interactions (\S\ref{sec:auto_extraction}). We validate our method, both quantitatively and qualitatively, finding that it correctly extracts the feedback to a reasonable degree.
We use our extraction method to obtain over $170k$ feedback samples from 1M non-annotated conversations. We release it as a dataset (\S\ref{sec:natural_feedback_dataset}).\footnote{Code and data are attached to the submission.}\cready{\footnote{Code and data: \url{https://github.com/shachardon/naturally_occurring_feedback}, \url{https://huggingface.co/datasets/shachardon/naturally_occurring_feedback}}}

To demonstrate the usefulness of the data, we use it to train a model to better align with human preferences. The resulting model demonstrates superior performance, outperforming the pretrained model in up to $79\%$ of the test cases (\S\ref{sec:training}).


\section{Background}

To compile a preference dataset, human annotators are asked to rank/score the generated responses of LLMs at the time of the interaction \citep{chiang2024chatbot}, or in retrospect \citep{bai2022training, pmlr-v162-ethayarajh22a}. 
To save this costly human effort, sometimes other models are doing the ranking \citep{cui2023ultrafeedback, lee2023rlaif, zhu2023judgelm} at the expense of introducing noise and biases \citep{zheng2024large}. 
For example, it was shown that LLMs tend to prefer longer responses regardless of their quality \citep{saito2023verbosity}. 
Although automatically extracted, naturally occurring feedback differs from model as a judge methods \citep{liu-etal-2023-g, zheng2023judging} as it is anchored in the human response, and therefore is less prone to ``hallucinations'' \citep{lewis2021retrievalaugmented} and biases \citep{saito2023verbosity}, and easier to explain and verify.

Another line of work collects data samples online during the interaction, by eliciting free-text feedback from the user. This feedback is then used in various ways for training \citep{shi2022life, jin2023data, scheurer2022training}. \citet{hancock-etal-2019-learning} suggested estimating user satisfaction and only if it is low, to elicit feedback from the users.

We focus on naturally occurring feedback, i.e., spontaneous unsolicited feedback.
When two humans talk, they do not score each other's responses nor explicitly ask for feedback (at least not often). 
Rather, the interlocutors actively signal their understanding and agreement through the use of verbal and visual responses, such as ``hmm'', ``yeah'' or facial expressions, head nods, etc. \citep{VRANJES201815, doi:10.1080/10904018.2010.508675}.

We show that also in a human-model textual conversation, such feedback signals exist. Finding them, ideally automatically, will allow us to extract freely annotated training examples. 
Extracting feedback from an endless stream of conversations \citep{donyehiya2024sharelmcollectionplugincontributing} has the potential to grow unboundedly, becoming a valuable complementary to other human feedback resource.

\section{The Discovery of Natural Feedback} \label{sec:natural_feedback}

We begin by defining a taxonomy for naturally occurring feedback. We then manually annotate conversations to account for the statistics of such feedback types in conversations. 


Throughout our discussion when we consider {\it feedback} we refer to (a part of) a human response that refers to (a part of) the last model's response.

\subsection{Feedback Taxonomy} \label{sec:feedback_taxonomy}

We define the following categories, split into four negative feedback categories and one positive:

\begin{enumerate}
    \item 
        Repeat or Rephrase (\textbf{rephrase}): The user repeats or rephrases their last response, explaining again what they wants. 
    \item 
        Make Aware with Correction (\textbf{aware + correct)}: The user points to the model that it was wrong, and provides information regarding the error/how to fix it. E.g., \textit{No, I wanted...}
    \item 
        Make Aware Without Correction (\textbf{aware -correct}): The user points to the model that it was wrong, without providing any additional information. E.g., \textit{That's incorrect}
    \item 
        Ask for Clarification (\textbf{clarify}): The user asks for additional resolution that was expected to be in the the previous response, but was missing. E.g., \textit{Was it like that?}
    \item 
        Positive Feedback (\textbf{positive}): The user confirms that the model did a good job, possibly thanking it. E.g., \textit{Thank you!}
\end{enumerate}

We now turn to motivating this set of categories. 
The two main design features are simplicity and 
text-anchoredness, i.e., the feedback should be directly and explicitly derived from the text, without requiring complex subjective interpretation.

Following this line, the feedback type that appears the most explicitly in the text is ``Positive Feedback''. Although we found it to be less common (see \S\ref{sec:natural_feedback_dataset}), positive feedback can usually be recognized at the vocabulary level. The user thanks the model (e.g., {\it ty}), says it did a good job (e.g., {\it great!}) or that it was right (e.g., {\it that's correct}).

The negative feedback cases, on the other hand, are much more diverse. There are vocabulary-level feedback cases ({\it that's wrong}), but also more semantically complex instances ({\it actually, I was asking about...}). 
Thus, using the feedback patterns from \citet{petrak-etal-2023-learning}, we break the negative feedback cases into finer categories to avoid too general a definition. Also, more detailed categories provide additional information that can be used later for better training/inference.

We found the ``Ask for Clarification'' category to be somewhat in the middle in terms of sentiment and feedback nature between the Positive Feedback and the rest of the negative categories. The user asks for more information or confirmation, indicating that the model's response was in the right direction, so not entirely wrong, but still provides some subtle feedback. This category is very common (see \S\ref{sec:auto_extraction}), and we expect these cases to be even more frequent as models improve.

Another distinction we found useful is between ``Make Aware with Correction'' and ``Make Aware without Correction''.
The first holds clear potential for training/inference, as the user provides information regarding the required fix. The latter is less useful, but still can be used as a strictly negative example (in contrast to Ask for Clarification).

``Repeat or Rephrase'' is unique compared to the other feedback forms as it does not leverage the model's ability to process multi-turn interactions.
Instead, the user ignores the previous response and rephrases again what they want, as if it was the beginning of the conversation. Assuming that the following model's response would be better, the two one-turn user-model interactions can be used as a preference pair for training. However, it is important to note that the context of the full conversation is crucial to recognize this feedback form. 

In their taxonomy, \citet{petrak-etal-2023-learning} also have an ``Ignore and Continue'' category, where the user ignores an error. We leave it out as it does not contain feedback, but rather implies that none was left despite an error. It is only meaningful when accompanied by an annotated error in the previous model response, which we do not have in our setting.



\begin{figure}[t]
\includegraphics[width=\columnwidth]{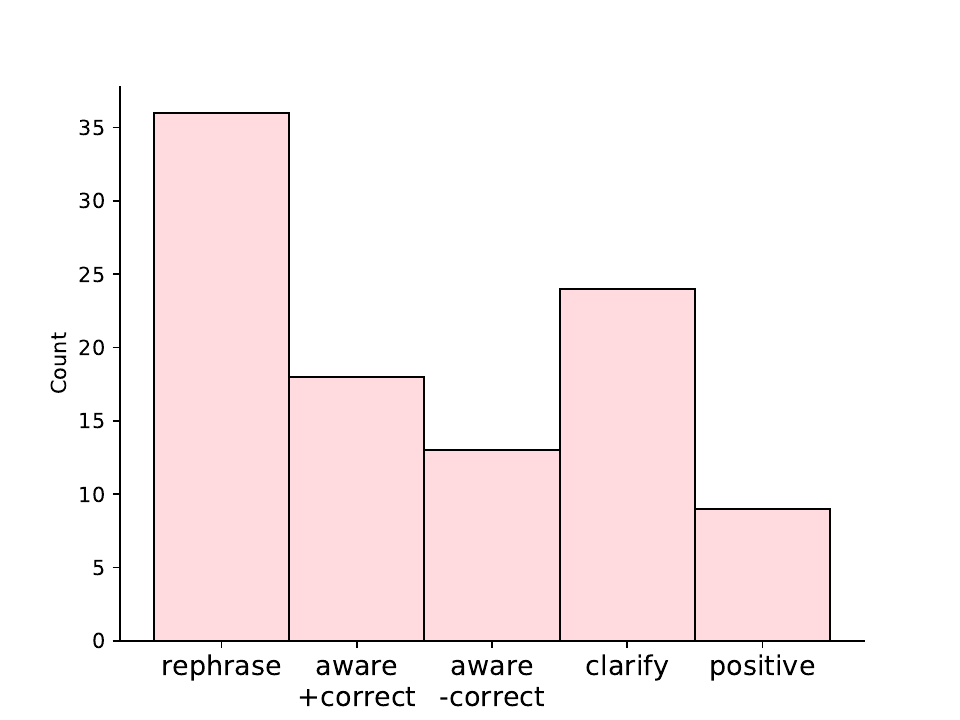}
\caption{The distribution of feedback categories for the first $300$ conversations in the dataset, as deemed by manual annotation. The most frequent categories are ``Repeat and Rephrase'' and ``Ask for Clarification''. There are only $9$ cases of ``Positive Feedback''.\label{fig:manual_categories}
}
\end{figure}

\subsection{Natural Feedback Commonly Occurs } \label{sec:manual_annotations}

To get an initial impression of the distribution of categories in this taxonomy, one of the authors manually annotated the first $300$  multi-turn conversations from the LMSYS-Chat-1M dataset \citep[see \S\ref{sec:implementation_details}]{zheng2023lmsyschat1m}.
After filtering out non-English conversations and offensive/unsettling conversations, we were left with $223$ conversations.
We find $77$ conversations with a total of $101$ feedback cases: $37$ Repeat or Rephrase, $18$ Make Aware with Correction, $13$ Make Aware without Correction, $24$ Ask for Clarification and $9$ Positive Feedback (see Fig.~\ref{fig:manual_categories}). The fact that $\sim30\%$ of conversations include feedback is an encouraging result. As the percentage is so high it is likely that simple methods would already suffice to extract notable amounts of feedback data.

To validate our manual annotation, we ask an in-house annotator to re-annotate the first $100$ conversations, of which $68$ pass the filtering. We get a Cohen's kappa of $0.65$ for the binary task of feedback recognition (see \S\ref{sec:extraction_eval}). Of the feedback cases that both annotators agreed upon, they also agreed on the category in $0.79$ of the cases.

\subsection{Prevalence Increases with Newer Models} \label{sec:up-to-date-feedback}

In the previous section we saw that current conversations contain a decent amount of feedback cases. Here, by comparing to older datasets, we try to asses what to expect in the future.

The state-of-the-art of LLMs advances rapidly \citep{open-llm-leaderboard}. In the interim, as models get better, users expect more; Users use the models for new scenarios that were not possible before \citep{zhao2024wildchat} and do so in a more natural way (except in extreme cases \cite{don-yehiya2023human}). 
We expect that with more fluent and diverse conversations, we will see more feedback.

We measure that empirically by comparing our annotation of current models to the annotation efforts of earlier models.
Out of the six datasets that were annotated by \citet{petrak-etal-2023-learning}, only the \textit{Self-Feeding Chatbot} dataset \citep{hancock-etal-2019-learning} is both human-model and open domain, and thus comparable. The Self-Feeding dataset was created in $2019$, and so is the model that was used to generate it.
Only $11$ feedback instances were found within a random sample of $100$ conversations. This is less than half the feedback frequency found in the newer LMSYS-Chat-1M dataset (omitting the positive feedback category as it was introduced by us). We note that there are $48$ annotated errors in the $100$ Self-Feeding dataset sample, and hence it is unlikely that it was a lack of errors that caused the users to give less feedback.

Our findings suggest that more than ever, naturally occurring feedback can serve as a valuable resource for feedback data. 
We believe in the future not only would models be better, continuing the above trend, but natural feedback itself may become a known resource, one which users expect the models to use (see \S\ref{sec:future}). 

\section{Automatically Extracting Feedback} \label{sec:auto_extraction}

Given that natural feedback is already present in current ongoing human-model conversations, we propose a method to automatically extract it.

Based on the five feedback categories (\S\ref{sec:feedback_taxonomy}), we instruct an LLM to recognize spans -- part of the human responses that contain feedback in a given conversation and classify them. We then use a Python script to parse the generated response and extract all the feedback instances. We discuss the implementation details next.

\begin{figure}[t]
\includegraphics[width=\columnwidth]{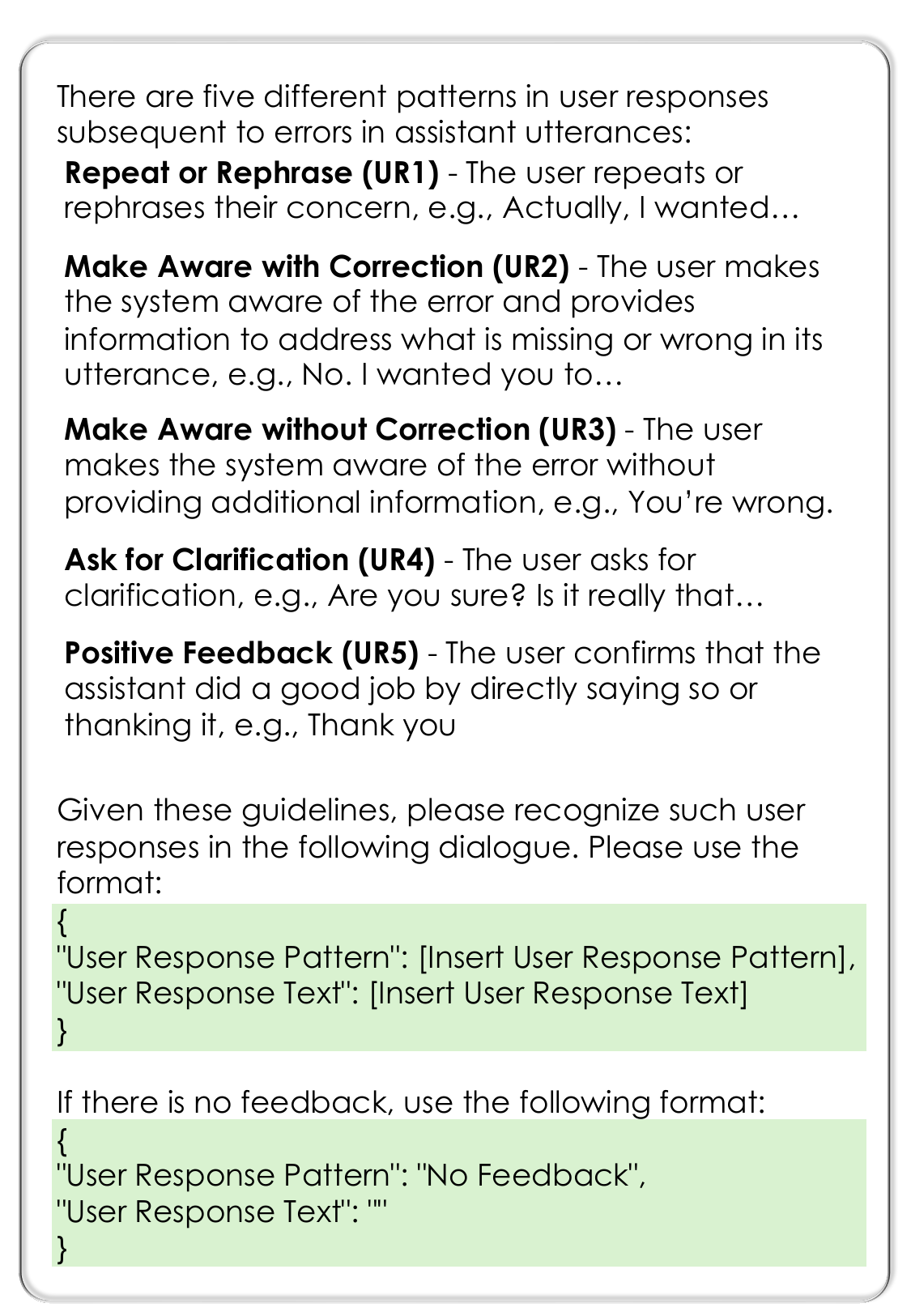}
\caption{Extraction Prompt. We describe the taxonomy and ask the model to output the categories and spans of human responses that contain feedback.
}
\label{fig:prompt}
\end{figure}

\subsection{Extraction Implementation Details} \label{sec:implementation_details}

\paragraph{Data.}
We use the LMSYS-Chat-1M dataset \citep{zheng2023lmsyschat1m}, a collection of real-world conversations with 25 state-of-the-art LLMs. We select this dataset for its size and variety of models and conversation topics. 
We filter out conversations with less than two turns, as there is no human feedback in a one turn conversation (one user query followed by one model response).

\paragraph{Model.}
We use \textit{Mixtral-8x7B-Instruct-v0.1} \citep{jiang2024mixtral} with 4-bit quantization (see App.~\S\ref{app:training_param}). During development, we also experimented with \textit{Yi-34B-Chat} \citep{ai2024yi} and \textit{GPT-3.5}, but found that Mixtral surpasses them. 

\paragraph{Prompt.}
After experimenting with a couple of versions, we found the prompt in App.~\ref{app:alternative_prompts} to perform best. 
One key point is asking the model to provide its output in JSON format, to ease the parsing. Using few-shot examples seems to confuse the model, probably due to the conversations length and the difficulty to separate different conversations. 

\paragraph{Parsing.}
If the generated text contains the prompt, we delete the prompt. We then extract all JSON objects and confirm they contain the ``User Response Pattern'' and ``User Response Text'' fields. For each of the JSON objects we verify that the ``User Response Text'' is indeed contained in one of the user responses and that the category is valid (one of the 5 possibilities). If any of these do not hold, we discard the example.

\begin{figure}[t]
\includegraphics[width=\columnwidth]{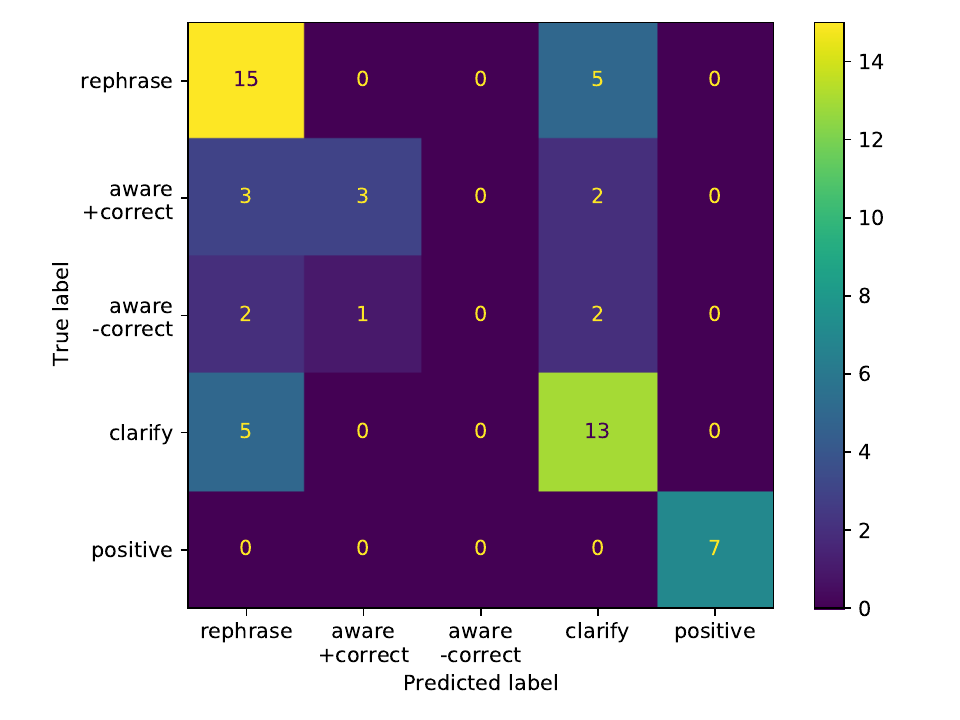}
\caption{Confusion Matrix for the Extracted Feedback. Out of the $101$ manually annotated feedback cases, our automatic method managed to find $58$, and to correctly classify to categories $38$. There is no confusion between ``Positive Feedback'' and the rest of the categories.
}
\label{fig:confusion}
\end{figure}

\begin{figure}[t]
\includegraphics[width=\columnwidth]{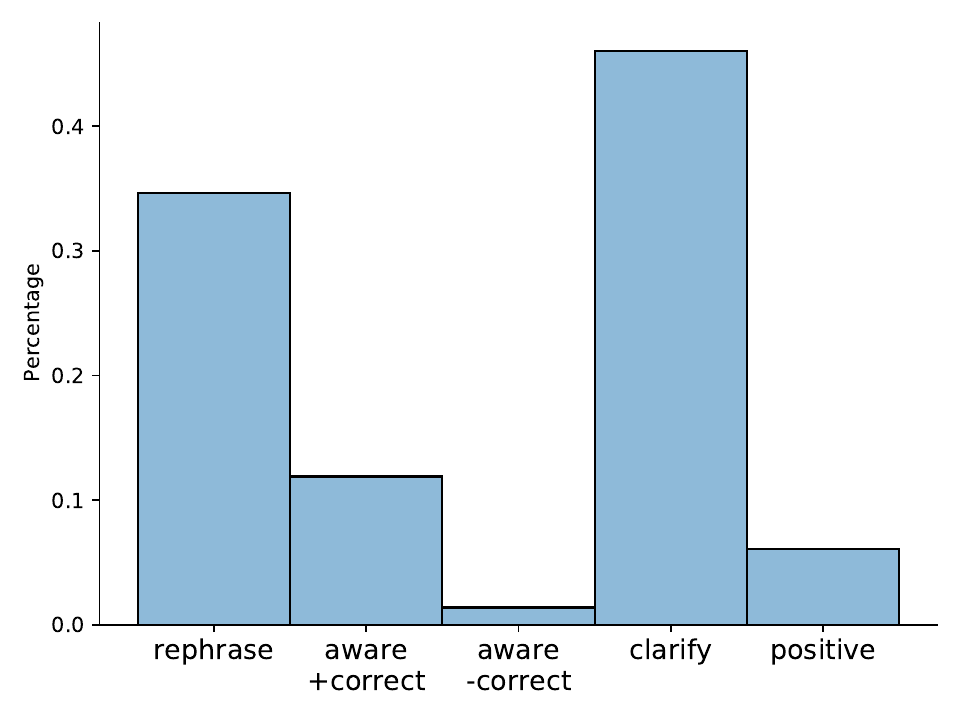}
\caption{Automatically extracted feedback distribution. The automatic and manual extraction (Fig.~\ref{fig:manual_categories}) agree on which  categories are more common: ``Ask For Clarification'' and ``Repeat or Rephrase''. ``Make Aware without Correction'' and ``Positive Feedback'' are the rarest.
}
\label{fig:automatic_categories}
\end{figure}

\subsection{Extraction Evaluation}\label{sec:extraction_eval}

To evaluate model performance in correctly identifying and classifying the naturally occurring feedback, we use the 300 manually annotated conversations from \S\ref{sec:manual_annotations}.
We analyze the two abilities separately; extraction of feedback span and classifying it into the correct category. A feedback span is considered correct if it is a sub-string of a manually annotated feedback span and is at least half as long.
We report both precision and recall.
We define the span-precision as the percentage of the correctly identified feedbacks out of the total number of identified feedbacks. Correspondingly, the span-recall is the ratio of those correctly identified feedbacks, but divided by the total number of manually annotated feedbacks. We define the category precision/recall as the number of feedbacks that were both identified and classified correctly to the right category, divided by the total number of identified/manually annotated feedbacks respectively.

While our manual annotation found $101$ feedback cases, our automatic method found $134$ feedback cases, out of which $58$ are correct.
Employing bootstrap with 1000 repetitions, this results to $0.43\pm0.05$ precision and $0.58\pm0.06$ recall. Taking the categories into account, we get $0.28\pm0.04$ precision and $0.38\pm0.06$ recall.
Fig~\ref{fig:confusion} presents the confusion matrix for the categories. 
We can see that ``Repeat and Rephrase'' and ``Make Aware with Correction'' are the most frequent categories for both manual annotation and the automatic method, and that the automatic method did not predict the ``Make Aware without Correction'' category at all.
If we reduce it to binary categories, i.e., positive/negative, we can see that there is no confusion between the positive category and the negative categories.


Examining the false positives, we see that many of them are debatable. For example, the user started by telling the model \textit{``I am interested to know how you work''}. The model then responded with \textit{``I am an AI language model that uses machine learning algorithms to understand and generate human-like text. I am trained on a large dataset of text... If you have any specific questions about how I work, feel free to ask!''}. The user then asked \textit{``Can I make you on my own computer?''} and our extraction method marked this text as a ``Ask for Clarification'' feedback case. We did not annotate this response as feedback as we considered it to be a new request and not a clarification of the previous one. However, we do see why it could be pointing to missing information in the original question. 

Given that also the false positives seem to encode a relevant signal, we hypothesize that although not perfectly retrieved, the extracted feedback would be beneficial for training. We validate it in \S\ref{sec:training}.

\subsection{The Natural Feedback Dataset} \label{sec:natural_feedback_dataset}

Using the manual annotation as a test set, and our extraction method to acquire more feedback, we create a large Natural Feedback Dataset.
We run the described extraction method on all 1M conversations of the LMSYS-Chat-1M dataset. After filtering out two turn conversations (see \S\ref{sec:implementation_details}), we are left with $334,319$ conversations. We apply our method and end up with $173,859$ feedback examples from $115,312$ different conversations. See Fig.~\ref{fig:automatic_categories} for the category distribution. In terms of positive/negative examples, we have about $15$ times more negative examples, similar to the ratio we had in the manual annotation (\S\ref{sec:manual_annotations}). Note that this ratio is not surprising, as correcting a model is potentially beneficial for the user (helping the model to help me), while thanking it is less practical.  
For more statistics and running the extraction method on another conversation dataset, see App.~\S\ref{app:dataset_stat}.

\section{Validating the Feedback Usefulness}\label{sec:training}

To demonstrate the usefulness of the extracted data, we use it to train LLMs and show the improvement.

Our data contains both positive and negative examples. We start by using the positive examples only, to finetune the models. We then present some initial results for preference training, with both positive and negative examples.

\subsection{Training Details} \label{sec:training_details}

We randomly split the positive examples to 80/20\% for train/val data, remaining with $8448$ training examples. 
We use three models in increasing sizes: \textit{EleutherAI/pythia-1.4b}, \textit{pythia-2.8b} \citep{biderman2023pythia}, and \textit{mistralai/Mistral-7B-v0.1} \citep{jiang2023mistral}. For more details see App.~\S\ref{app:training_param}.

\subsection{Model Performance Evaluation}\label{sec:evaluation}

To measure the improvement of the models given the new data, we use the validation split of the OpenAssistant dataset \citep{kopf2024openassistant}. We generate the last response with both our trained models and the corresponding pretrained models. See App.~\S\ref{app:training_param}.


\paragraph{Human Evaluation.}
We perform human evaluation of the model outputs to acquire a reliable evaluation.
An in-house human annotator was asked to rate not consistently ordered pairs of $100$ model responses for each of the models, without knowing what model created which response (the pretrained baseline or the finetuned version). Our trained models won $69\%$ / $81.5\%$ / $77\%$ over their corresponding pretrained versions.

\paragraph{Evaluation by Open Models.}
In addition to manual evaluation, we perform automatic evaluation, which allows more flexibility in the analysis.
To prioritize replicable science, we explore the usability of open models as evaluators in our scenario.
The RewardBench leaderboard \citep{lambert2024rewardbench} evaluates the capabilities of models in the task of rating model responses, and its top models outperform some closed models that are frequently used as judges. Based on the leaderboard, we take \textit{openbmb/Eurus-RM-7b} \citep{yuan2024advancing} and \href{https://huggingface.co/sfairXC/FsfairX-LLaMA3-RM-v0.1}{sfairXC/FsfairX-LLaMA3-RM-v0.1} (leaders of the leaderboard, when conducting these experiments), and compare the pretrained to the finetuned models responses.
However, we were disappointed to find that these models do a poor job comparing the outputs of the smaller models. Eurus-RM-7b reported $31\%$ and $38\%$ wins for the 1.4B and 2.8B models, and FsfairX-LLaMA3-RM-v0.1 reported $48\%$ and $60\%$.
We assume this is due to the distribution of the data they were trained on, which only represents stronger models.
The results for the 7B model on the other hand are comparable to those we got in the human and GPT evaluation. Eurus-RM-7b and FsfairX-LLaMA3-RM-v0.1 reported $70\%$ and $72\%$ wins for the trained model respectively.

\paragraph{GPT as a Judge.}
To complete the picture, we run a GPT-4 as a judge evaluation \citep{zheng2023judging}. 
We use the RewardBench\citep{lambert2024rewardbench} implementation to instruct GPT-4 to compare response pairs.
Our trained models won $65\%$ / $74\%$ / $78\%$ over their corresponding pretrained versions. 

\paragraph{}
Conducting a binominal test on these results, we find that all above reported results are significant with $p<10e^{-9}$.
We conclude that our automatically extracted training data is indeed beneficial. Training on about $8k$ positive examples yields a significant improvement for all our tested model sizes, more as the model is larger. See App.~\S\ref{app:results_table} for the aggregated results table.

\subsection{Random Chats Baseline}\label{sec:random_chats}
As an additional baseline, we replace our extracted positive examples with a random sample of chat examples from the LMSYS-Chat-1M dataset of the same size. These examples are not necessarily positive, but they are in a chat format and of relatively well-performing models and therefore might be useful for knowledge distillation nonetheless \citep{honovich-etal-2023-unnatural}. We want to test whether training on our extracted data has any advantage over this randomly sampled data. We finetune the 7B model on them, and evaluate their performance. Eurus-RM-7b reports $64\%$ wins, FsfairX-LLaMA3-RM-v0.1 reports $68\%$ wins, and GPT-4 reports $75\%$ wins, all outperformed by the model we trained on the extracted positive examples. This strengthens our conclusion that our extracted data is beneficial.


\subsection{Preference Training}\label{sec:preference}

So far, we have shown promising results for finetuning. While finetuning is a performant way to use feedback, it only trains on positive examples (see \S\ref{sec:training_details}). To test the benefits of the negative examples we try other training methods.

\paragraph{Experimental Setup.}
As mentioned in \S\ref{sec:natural_feedback_dataset}, there are many more negative examples than positive ones.
To balance this, we use only the "Make Aware with Correction" and "Make Aware without Correction" categories, and on top of that we down-sample. We chose these categories as we assume their 'negative' signal is the strongest.  
For the hyperparameters, see App.~\S\ref{app:training_param}.
The positive and negative examples are not given in pairs that share a prompt and have varying quality degrees. We hence use KTO \citep{ethayarajh2024kto}, which improves over DPO objective \citep{rafailov2024direct}, while also handling non-paired preference data.\footnote{The use of DPO might be possible for feedbacks of the rephrase category, but as these account for one category only, focusing on them is less relevant for our purposes here, of demonstrating the usefulness of the extracted feedbacks.} 



We run this experiment with the 7B model only, as preference training is not beneficial for smaller models \citep{ethayarajh2024kto}. We start from the previously finetuned model.

\paragraph{Results.}
All tuned models score overwhelmingly better than the pretrained.
Eurus-RM-7b reports $74\%$ wins, FsfairX-LLaMA3-RM-v0.1  reports $75\%$ wins, and GPT-4 reports $79\%$ wins. Those are also somewhat better than those we had for the finetuned model ($1$-$3$ points improvement). We conclude that our negative data (or at least some of the categories) is indeed useful.

\section{Ablation Experiments}\label{sec:analysis}

We analyze different aspects of our extraction method, including our choice of taxonomy, and other frequently used prompting techniques.

\subsection{Taxonomy Effect on the Extraction} \label{sec:categories_choice}

Here we examine the effect of the feedback taxonomy on the success of the model in accurately extracting the feedback spans from the conversations.
We examine several taxonomy alternatives. We evaluate each by calculating the precision and recall relative to the 300 manually annotated conversations from \S\ref{sec:manual_annotations}.

\subsubsection{No Categories}

Our taxonomy introduces $5$ different feedback categories (see \S\ref{sec:feedback_taxonomy}). Here we examine whether there is even a need for any taxonomy at all.

We change our prompt such that it will not contain any category definition. We instruct the model to recognize spans of text that are informitive as to the satisfaction of the user, and rate them on a scale of 1-5. See App.~\S\ref{app:alternative_prompts} for the prompt.

Running the model in this setting, the model found $693$(!) text spans, while not even one of them matches the manually annotated feedback examples. Manually looking at a few of them, it seems that the model fails miserably at identifying relevant text spans. For example, it often suggests seeing the original user's requests as an indication of user satisfaction (e.g., \textit{``Show me how to implement a toy version of a relational database. Begin by writing a toy query planner that convert SQL...''}), which is of course not a valid user feedback as it precedes the model response.

We conclude that an overly general extraction prompt is harder for the model to handle, and that a detailed taxonomy helps automatic extraction.

\subsubsection{Limited Categorization}

We examine the effect of using fewer feedback categories on the extraction process. This followed the hypothesis that focusing on a smaller set of categories would allow for better precision.

We limit ourselves to the ``Repeat and Rephrase'' and ``Positive Feedback'' categories, as we recognize that they are both relatively easier for the model to distinguish from the other categories (see \ref{fig:confusion}). We instruct the model to extract these two types for feedback only. See App.~\S\ref{app:alternative_prompts} for the prompt.

For the ``Positive Feedback'' category, the model manages to achieve $0.5$ precision for both span and category precision, i.e., all positive cases that the model found were classified correctly.

For the ``Repeat or Rephrase'' category, the model managed to achieve $0.43$ text-span precision and $0.17$ category precision. Given that there are only two possible categories, there is a relatively large gap between the span and category precision.
Looking at the extracted examples, we see that the model tends to invent new categories, for example ``Asking for Assistance'' or ``Ask for Examples''. 

We conclude that focusing on fewer feedback categories is not necessarily easier for the model.


\subsection{Confidence Level}
We examine the usefulness of asking the model to generate a ``confidence level'' value, to better filter the extracted feedback samples such that we will get a higher precision score.
To do so, in addition to the ``User Response Pattern'' and ``User Response Text'' fields, we instruct the model to provide a ``Confidence Level (1-5)'' field. See App.~\S\ref{app:alternative_prompts}.

Looking at the distribution of confidence scores the model assigned, we find that over $96\%$ of the feedback cases received a $5$ score. The other $4\%$ are mostly ``No Feedback'' or hallucinations that are automatically removed at the parsing stage (see \S\ref{sec:implementation_details}).
We conclude that this method is ineffective.



\section{Related Work}\label{sec:related_work}

In addition to the LMSYS-Chat-1M dataset which we used due to its size and inclusion of multiple models, there are other recent datasets such as \textit{Collective Cognition}\footnote{\url{https://huggingface.co/CollectiveCognition}}, \textit{PRISM} \citep{kirk2024PRISM}, and \textit{WildChat} \citep{zhao2024wildchat}. We use the last to replicate the feedback statistics, see App.~\ref{app:dataset_stat}.

\citet{petrak-etal-2023-learning} investigated the types of errors and user responses in $6$ different datasets. Although we adopt and modify their feedback taxonomy, we take two steps further. We focus on user responses and extract them automatically, and we show the importance of using up-to-date conversation data (\S\ref{sec:up-to-date-feedback}), contrary to their conclusion.

We opted for KTO to train a model on our data, but there are many more options for training on non-positive examples \citep{christiano2017deep}. \citet{NEURIPS2022_b1efde53} and \citet{shi2022life} suggested to create possible corrections for negative examples and train on them. Other methods use pairs of positive and negative examples and train on them both to predict their scores \citep{liu2023chain}.
\citet{peng2024pragmaticfeaturepreferenceslearning} and \citet{wu2024fine} showed the possible gain of fine-grained feedback.

We want to emphasize the difference between works that use \emph{natural language feedback} \citep{yan-etal-2023-learning, enhancing_performance, sreedhar-etal-2020-learning} and ours. In their settings, the user is asked to provide free-text feedback (unlike binary feedback or other close forms). We take another step further and look for feedback that spontaneously occurs in the conversations, without explicitly asking the user to provide it.

Perhaps the most closely related work to ours is \citep{lin2024interpretable}. However, our approach differs in several key aspects: focusing on feedback retrieval over binary classification, using extrinsic rather than intrinsic evaluation, and employing an open model instead of a closed one.

Another set of related works are ones creating synthetic data from datasets \citep{yehudai2024genie} or augmenting user feedback \citep{sudalairaj2024lab}. We note their similarity in better utilizing human effort for creating data samples for training. Those, however, differ from our work in the problem they address. Such efforts rely on boosting existing training signals, whether found in the human explicit annotations, the model, or both. In contrast, our approach aims to identify signals in a scalable fashion. In fact, the output of our method can be used as their input \citep{bartolome2024distilabel}.

\section{Discussion and Future Work}\label{sec:future}

This paper advocates the use of naturally occurring feedback and introduces a method to extract it. We find that naturally occurring feedback is common in human-model chats. We use our method to extract over $170k$ feedback samples and train models on them, demonstrating their usefulness.  


We observed in \S\ref{sec:up-to-date-feedback} that newer conversation data tends to contain more naturally occurring feedback. \citet{buschmeier2018communicative} showed the importance of ``listener feedback'' (subtle verbal signals, head gestures, and facial expressions) for communication. They showed that this feedback encourages humans interacting with a model to provide more feedback by themselves, and to rate the conversation as more helpful. 
Therefore, we expect voice assistant data to contain even more feedback.

Another interesting line of future work is the incorporation of feedback into chats in real-time, with interactive reinforcement learning for example, or at least in a manner that would directly affect future user conversations, making giving feedback more beneficial for the user.

\section*{Limitations}\label{sec:limitations}

Although the original LMSYS-Chat-1M dataset contains some non-English conversations, we filter those out during evaluation, as our annotators are not familiar with these languages. 

Our automatic extraction method can be improved further to achieve better precision and recall (e.g., with better models, prompts, or a more sophisticated extraction algorithm). We believe that the fact that even the current relatively low precision data managed to achieve good training results underscores the importance and potential of naturally occurring feedback.  
With the abundance of data, future work might seek better precision or keep high recall depending on their goals. One could train on a lot of low quality data, focus on specific subsets of interest (e.g., a domain) or focus on quality annotation throwing a lot and still no ending up wanting, each requiring different precision-recall tradeoffs.


We used GPT as a judge and other models for evaluating the trained model. This approach is both costly and known to have biases \citep[e.g.,][]{dubois2024length, panickssery2024llm}. Therefore, we use it only to complement the human evaluation and the open-models evaluation. We would like to emphasize that our extraction method itself does not use GPT or any proprietary models.

Another limitation of our evaluation method is that it does not allow absolute results, but instead compares the model at hand to another model. This raises the question of which model should be used for the comparison. Here, as in previous work (e.g., \cite{ethayarajh2024kto}), we selected one reference model for all our experiments, to allow comparison between the different settings. We used the pretrained model, as it was the best fit for the main experiment (pretrain vs. pretrain + finetune on our positive data).


\section*{Ethics Statement} \label{sec:ethics}

This work has been approved by the IRB of our institution.
We abide by the terms and conditions of the LMSYS-Chat-1M dataset (see the license here \footnote{\url{https://huggingface.co/datasets/lmsys/lmsys-chat-1m\#lmsys-chat-1m-dataset-license-agreement}}.
As mentioned by the LMSYS-Chat-1M dataset authors, the LMSYS-Chat-1M dataset contains unsafe conversations that may be perceived as offensive or unsettling. The provided OpenAI moderation API tag can be used to filter it. We informed our annotators of this and instructed them to skip these conversations.

\cready{
\section*{Acknowledgments and Thanks}
We thank John (autoMeta) Cook, Ramon Astudillo and Ben Burtenshaw for the deep discussions that helped us converge with our thoughts.
}
\bibliography{acl_latex}

\appendix

\section{Models and Parameters}
\label{app:training_param}
All the models we used were released with a apache-2.0 license, except to llama3 which is released with the llama3 license and OpenAI models with their own terms of use.

To run our extraction process, we run the model with 0.2 temperature, 256 maximum new tokens, top-p 0.95, and 1.0 repetition penalty. 
Overall, the model processed approximately one conversation per 10 seconds on an NVIDIA 40A GPU. 

We use a learning rate of $5e-7$ and RMSprop optimizer.
We use NVIDIA RTX 6000 for the 1.4B model, NVIDIA A40 for the 2.8B and 2 A40 for the 7B models. The pythia models were chosen for their sizes, and the Mistral for its popularity in the 7B category.
To fit our GPUs we restrict the maximum input length to 1024, and accumulate gradients to achieve a batch size of $32$. We run training for up to $20$ epochs, and select the best model according to its performance on the validation set.

For the KTO training, we use the same hyperparameters as in the finetuning experiment, and take the ones specific to KTO from an existing KTO implementation \footnote{\url{https://github.com/ContextualAI/HALOs}}.

Training each model took up to five days, depending on its size and the GPU used.
 
To evaluate the models, we use the same generation parameters as above: 0.2 temperature, 256 max new tokens, 0.95 top p, and 1.0 repetition penalty.

We used NVIDIA RTX 6000 for both generating the outputs and for running the open rewards models. Generating the outputs took up to two days for the 7B models and much less for the smaller ones. Running the rewards models took up to $15$ minutes for each. Using GPT-4 as a judge cost us $70\$$.

\section{Dataset Statistics}\label{app:dataset_stat}
We examine the statistics of the conversations that were found
to contain feedback (\S\ref{sec:natural_feedback_dataset}).
The average number of turns in a conversation in the LMSYS-Chat-1M dataset is $2$, while the average number of turns in a conversation that contains feedback is $5.5$. This is not surprising as the minimum number of turns in conversation that contains feedback is $2$ as user feedback can appear after at least one model response, i.e., starting from the second turn only.
The average feedback turn is $3.1$, and the average length of the feedback span is $52.5$ tokens.

To show that both our feedback taxonomy and extraction method are general, we run the extraction process on another dataset, WildChat (see \S\ref{sec:related_work}).

When running the extraction method on the first 1000 multi-turn conversations of the WildChat dataset, we found $485$ conversations that contain feedback. To compare, for the first 1000 multi-turn conversations of the LMSYS-Chat-1M dataset we found $454$ conversations. Thus, we conclude that the feedback distribution is consistent across other now-days feedback datasets.

\section{Annotators Instructions}
Here we describe the annotators guidelines.

\subsection{Manually Feedback Annotation}
For the feedback annotation task, the annotator was given the following guidelines:
\paragraph{}
\texttt{There are five different patterns in user responses subsequent to errors in assistant utterances:
\paragraph{Repeat or Rephrase (UR1) -} The user repeats or rephrases their concern, explaining again what they want.
\paragraph{Make Aware with Correction (UR2) -} The user points to the model that it was wrong, and provides information regarding the error/how to fix it. "No, I wanted..."
\paragraph{Make Aware without Correction (UR3) -} The user points to the model that it was wrong, without providing any additional information. "That's incorrect"
\paragraph{Ask for Clarification (UR4) -} The user asks for additional resolution that was expected to be in the the previous response, but was missing. "Was it like that?”, “Can you provide a code solution for this?” 
\paragraph{Positive Feedback (UR5) -} The user confirms that the model did a good job, possibly thanking it. "Thank you!", “Great!”}\\

\texttt{Given these guidelines, please recognize such user responses in the following dialogue.}

\texttt{Note that there might be conversations that contain unsafe, offensive or unsettling content. If you encounter such a conversation, please skip it.
To speed up the annotation, you can start by going through the user's responses only, and then if there is a feedback candidate, look at the entire conversation context.}

\subsection{Human Evaluation Annotation}
For the human evaluation annotation task we used the following guidelines:\\

\texttt{Each example includes a prompt, and two possible assistant responses: response\_a and response\_b.
Your task is to choose the better response (‘A’ or ‘B’).
If the responses seems about the same, or you are unsure about which one to pick, pick ‘TIE’.}

\texttt{Note that some of the prompts already contain some assistant responses, as they are the beginning of a multi-trun conversation. In that case, your task is to choose the better next assistant response.}\\

\noindent\texttt{Example:}\\

\noindent\texttt{prompt:}\\ 

\noindent\texttt{\# user: Describe the impact of climate change on coastal cities.}\\

\noindent\texttt{\# assistant:}\\

\noindent\texttt{response\_a:}\\

\noindent\texttt{\# user: Describe the impact of climate change on coastal cities.}\\

\noindent\texttt{\# assistant: Climate change leads to rising sea levels, which can cause flooding in coastal cities. It also increases the frequency of extreme weather events, damaging infrastructure and displacing residents. Efforts to mitigate these impacts include building sea walls and developing better drainage systems.}\\

\noindent\texttt{response\_b:}\\

\noindent\texttt{\# user: Describe the impact of climate change on coastal cities.}\\

\noindent\texttt{\# assistant: Climate change affects coastal cities by causing sea levels to rise and increasing the frequency of severe storms. These changes result in frequent flooding, infrastructure damage, and forced relocation of communities. Strategies to combat these effects include constructing barriers like sea walls, enhancing urban drainage, and implementing comprehensive coastal management plans.}\\

\noindent\texttt{Answer:}\\

\noindent\texttt{‘B’}

\section{Alternative Extraction Prompts} \label{app:alternative_prompts}

\begin{figure}[t]
\includegraphics[width=\columnwidth]{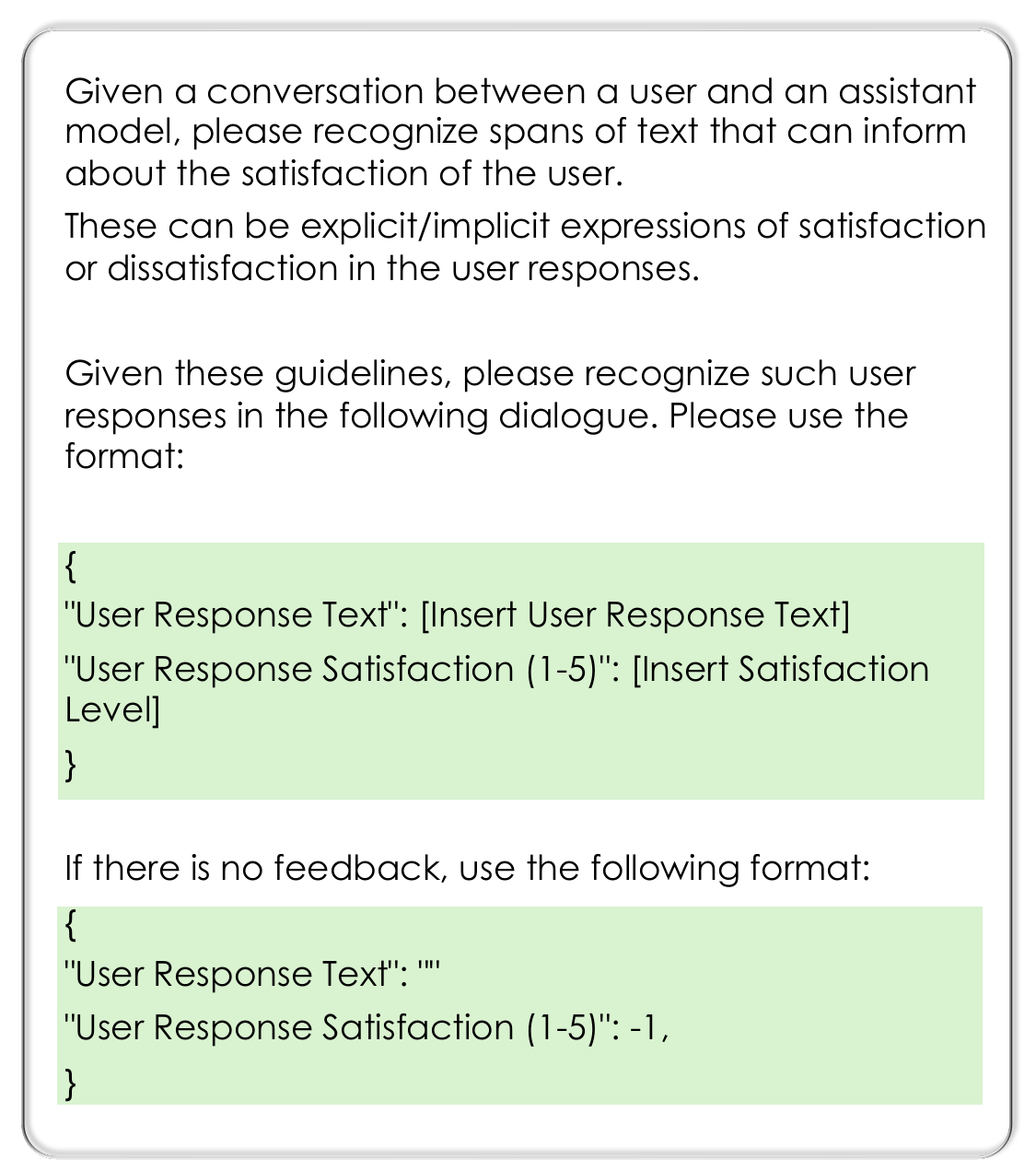}
\caption{Extraction Prompt for the No Categories setting.
}
\label{fig:prompt_co_cat}
\end{figure}

\begin{figure}[t]
\includegraphics[width=\columnwidth]{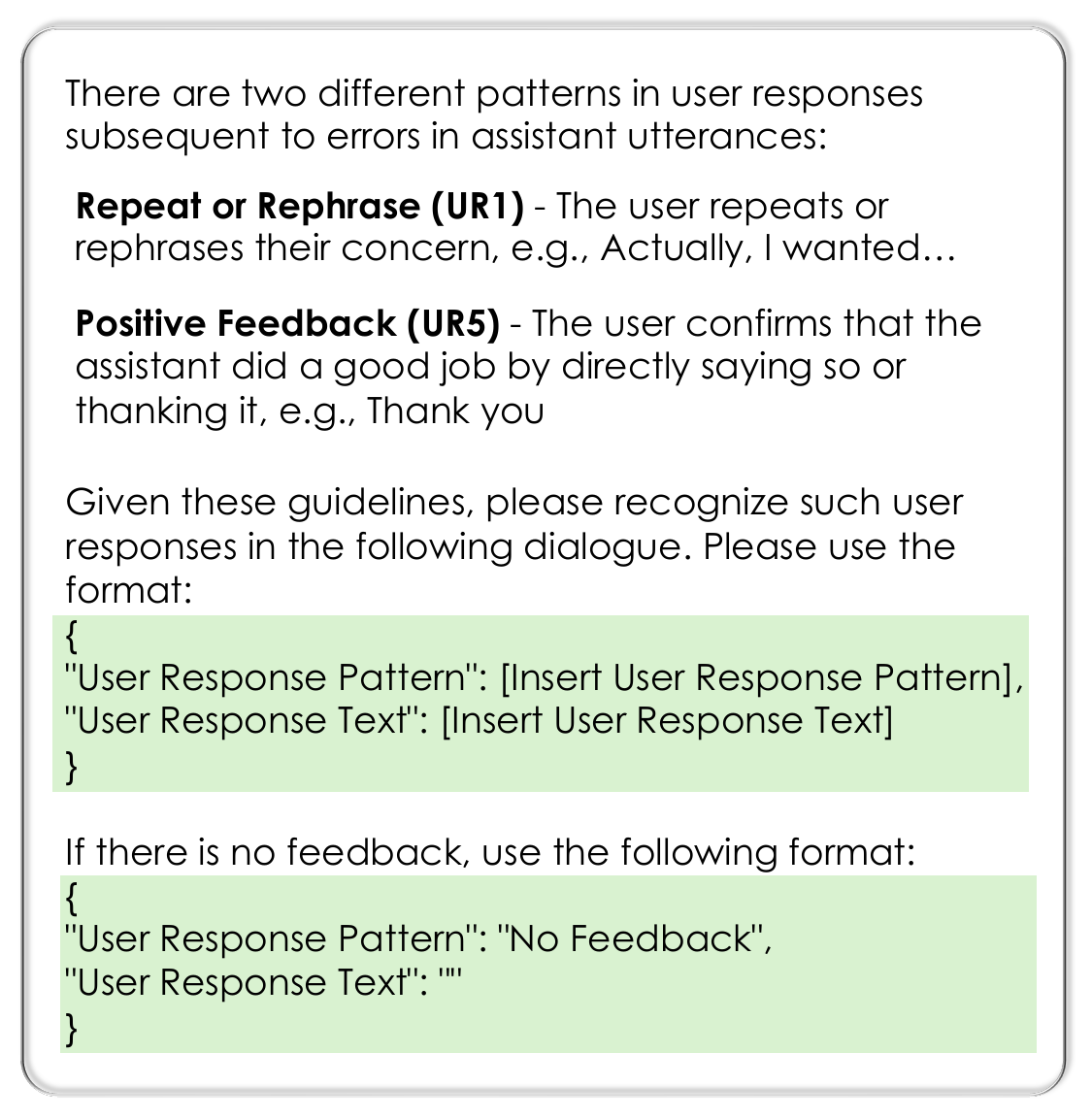}
\caption{Extraction Prompt for the Limiting Categories setting.
}
\label{fig:prompt_two_cat}
\end{figure}

\begin{figure}[t]
\includegraphics[width=\columnwidth]{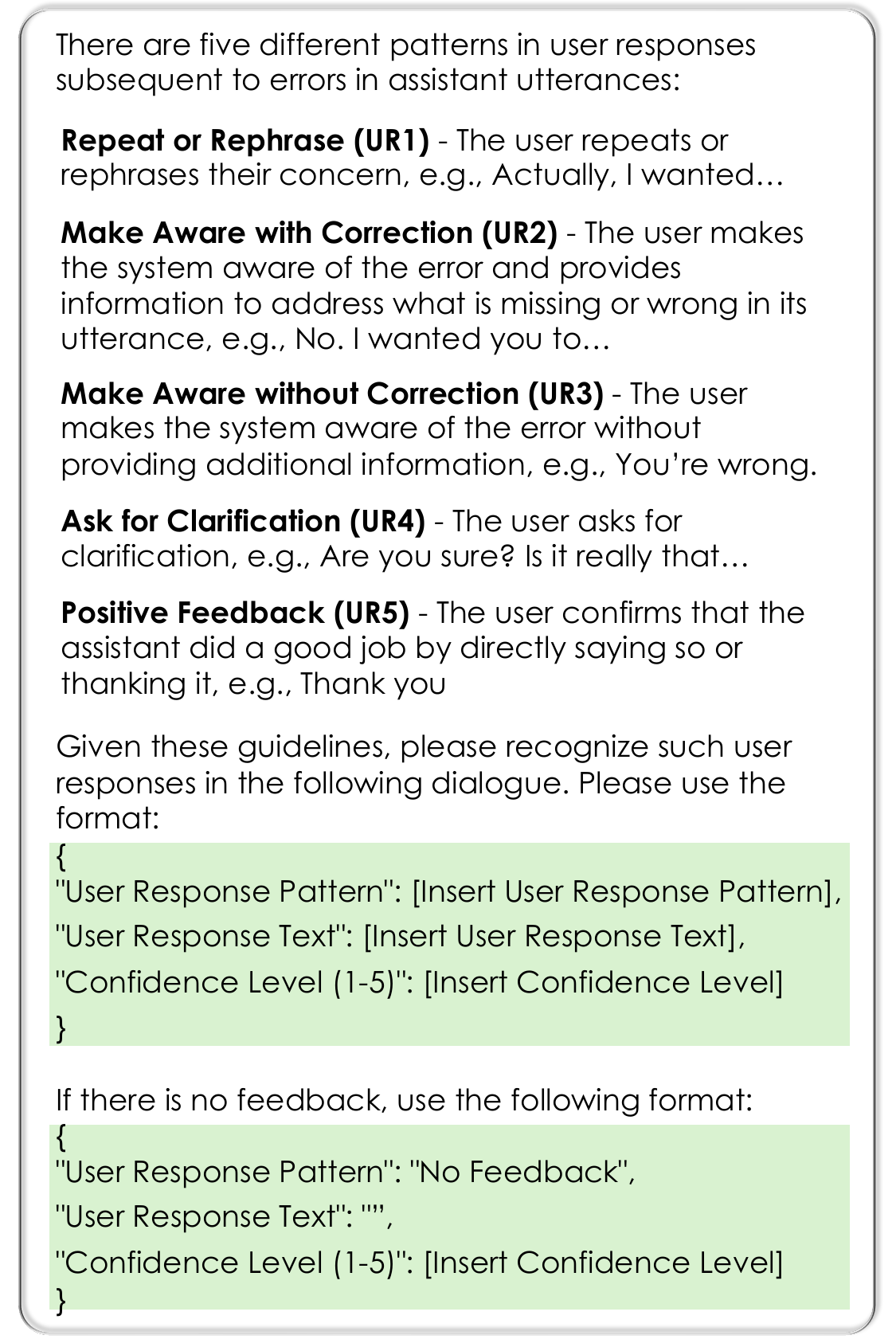}
\caption{Extraction Prompt for the Confidence Level setting.
}
\label{fig:prompt_confidence}
\end{figure}

For the ``No Category'' setting, we use the prompt in Fig.~\ref{fig:prompt_co_cat}.

For the ``Limiting Categories'' setting, we use the prompt in Fig.~\ref{fig:prompt_two_cat}.

For the Confidence Level setting, we use the prompt in Fig.~\ref{fig:prompt_confidence}.


\begin{table*}[h!]
\centering
\begin{tabular}{|l|c|c|c|c|c|}
\hline
\textbf{Training Method} & \textbf{Model Size} & \textbf{Human} & \textbf{Open Models -} & \textbf{Open Models -} & \textbf{GPT-4} \\ 
& & \textbf{Evaluation} & \textbf{Eurus-RM-7b} & \textbf{FsfairX-LLaMA3} & \textbf{Evaluation} \\ \hline
\multirow{3}{*}{\textbf{Random Chats}} & 1.4B & N/A & N/A & N/A & N/A \\ 
\textbf{} & 2.8B & N/A & N/A & N/A & N/A \\ 
 & 7B & N/A & 64\% & 68\% & 75\% \\ \hline
\multirow{3}{*}{\textbf{Finetuned}} & 1.4B & 69\% & 31\% & 48\% & 65\% \\ 
 & 2.8B & \textbf{81.5\%} & 38\% & 60\% & 74\% \\ 
 & 7B & 77\% & 70\% & 72\% & 78\% \\ \hline
\textbf{Finetuned +} & 7B & N/A & \textbf{74\%} & \textbf{75\%} & \textbf{79\%} \\ 
\textbf{Preference} & & & & & \\ \hline
\end{tabular}
\caption{Evaluation results for different training methods and model sizes, as described in sections \S\ref{sec:evaluation}, \S\ref{sec:random_chats} and \S\ref{sec:preference}. For each column the best win-rate is in bold. The best setting is Finetuned + Preference Training. Except for the open model evaluation of the smaller models (see \S\ref{sec:evaluation} for the discussion), the three evaluation methods (human evaluation, open models, GPT as a Judge) seem consistent.}
\label{table:results}
\end{table*}

\section{Results Table}\label{app:results_table}
Table ~\ref{table:results} aggregates the results from \S\ref{sec:evaluation}, \S\ref{sec:random_chats} and \S\ref{sec:preference}.

\section{Ai Assistants In Research Or Writing}
We used copilot for writing code scripts, and also used Chat-GPT a little for sentence rephrasing.

\end{document}